\documentclass[letterpaper, 10 pt, conference]{ieeeconf}  

\IEEEoverridecommandlockouts                              

\usepackage{graphicx} 
\usepackage{tabularx}
\usepackage{stfloats}

\usepackage{cite}
\usepackage{algorithm}
\usepackage{algorithmic}
\usepackage{amsfonts}
\usepackage{amsmath}
\usepackage{array,booktabs,arydshln}
\usepackage{bbm}
\usepackage{subfig}
\usepackage{booktabs}
\usepackage{float}
\usepackage{xcolor}
\usepackage{flushend}

\definecolor{green}{RGB}{11,155,13}

\title{\LARGE \bf
Grounded Curriculum Learning}

\author{Linji Wang$^{1}$, Zifan Xu$^{2}$, Peter Stone$^{2, 3}$, and Xuesu Xiao$^{1}$ 
\thanks{
        $^{1}$Department of Computer Science, George Mason University {\tt\small \{lwang44, xiao\}@gmu.edu}
        $^{2}$Department of Computer Science, The University of Texas at Austin {\tt\small zfxu@utexas.edu, pstone@cs.utexas.edu}
        $^{3}$Sony AI
        }%
}

\begin{document}

\maketitle
\thispagestyle{empty}
\pagestyle{empty}

\begin{abstract}

The high cost of real-world data for robotics Reinforcement Learning (RL) leads to the wide usage of simulators. 
Despite extensive work on building better dynamics models for simulators to match with the real world, there is another, often-overlooked mismatch between simulations and the real world, namely the distribution of available training tasks.
Such a mismatch is further exacerbated by existing curriculum learning techniques, which automatically vary the simulation task distribution without considering its relevance to the real world. 
Considering these challenges, we posit that curriculum learning for robotics RL needs to be \emph{grounded} in real-world task distributions.
To this end, we propose Grounded Curriculum Learning (GCL), which aligns the simulated task distribution in the curriculum with the real world, as well as explicitly considers what tasks have been given to the robot and how the robot has performed in the past. 
We validate GCL using the BARN dataset on complex navigation tasks, achieving a 6.8\%  and 6.5\% higher success rate compared to a state-of-the-art CL method and a curriculum designed by human experts, respectively. These results show that GCL can enhance learning efficiency and navigation performance by grounding the simulation task distribution in the real world within an adaptive curriculum. 
\end{abstract}

\section{Introduction}

\label{sec:intro}
Reinforcement learning (RL) has become a powerful tool that enables robots to learn complex behaviors through trial-and-error interactions with their environments~\cite{kober2013rl_robotics}. However, applying RL to real-world robotic tasks presents significant challenges. The trial-and-error process often requires a vast amount of data, which is difficult and expensive to collect in real-world settings~\cite{dulac2019challenges}. As a result, simulators have become widely used to generate training data in a more controlled and cost-effective manner.

While much work has been done to build simulators with better dynamics models that more closely match the physical world~\cite{zhao2020sim}, there is another, often-overlooked mismatch between simulations and the real world, namely the distribution of available training tasks. Specifically, the tasks generated in simulators may differ in complexity, variability, and structure compared to those that robots encounter after simulated training during deployment in real environments~\cite{kormushev2013reinforcement, xu2022learning}. This mismatch can hinder the generalization and performance of RL agents when transitioning from simulated environments to real-world tasks.

This problem is further exacerbated by existing Curriculum Learning (CL) techniques, which automatically vary the simulation task distribution to facilitate learning~\cite{bengio2009curriculum}. Recent approaches, such as PAIRED~\cite{dennis2020emergent} and CLUTR~\cite{azad2023clutr}, have demonstrated improved generalization by using teacher agents to generate a curriculum of increasingly complex tasks. However, these methods typically focus on optimizing the simulation curriculum without considering its relevance to the real world. Consequently, RL agents may be trained on simulation tasks that are not representative of real-world conditions, leading to poor performance upon deployment.
\begin{figure}[t]
    \centering
    \includegraphics[width=\columnwidth]{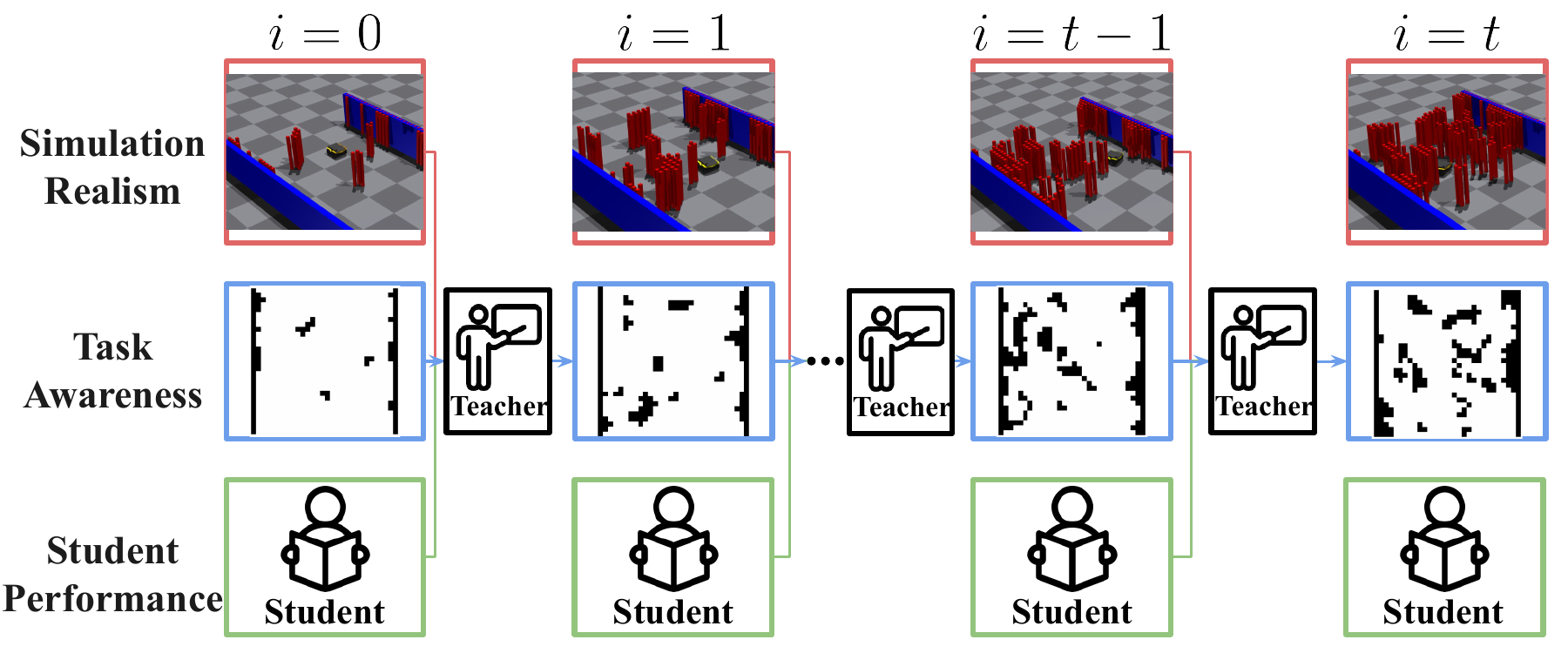}
    \caption{
    Considering simulation realism, task awareness, and student performance, GCL grounds its curriculum in real-world task distribution and creates an adaptive sequence of BARN navigation tasks with properly increasing difficulty. 
    }
    \label{fig:overview}
    \vspace{-15pt}
\end{figure}

Considering these challenges, we posit that curriculum learning for robotics RL needs to be \emph{grounded} in real-world task distributions. Grounding the curriculum in real-world tasks ensures that the learning process remains relevant and that the trained policies are more likely to generalize effectively when deployed in real-world environments.
To this end, we introduce Grounded Curriculum Learning (GCL), a framework designed to align the simulated task distribution with real-world tasks to assure real-world generalization while improving learning efficiency with an adaptive curriculum.   
GCL achieves this by considering three key aspects:
(1) simulation realism: aligning the simulated task distribution in the curriculum with the real world, 
(2) task awareness: tracking the sequence of tasks given to the robot, and
(3) student performance: monitoring the student's performance across previous tasks in the curriculum (Fig.~\ref{fig:overview}).

We validate GCL using the Benchmark Autonomous Robot Navigation (BARN) dataset, a standard testbed for evaluating robotic navigation performance in complex and highly constrained environments~\cite{perille2020benchmarking, xu2023benchmarking}. Our experiments demonstrate that GCL achieves a 6.8\%  and 6.5\% higher success rate compared to a state-of-the-art CL method and a curriculum designed by human experts, respectively. These results highlight GCL's ability to enhance both learning efficiency and navigation performance by grounding the simulation task distribution in the real world within an adaptive curriculum.

\section{Related Work}
\label{sec::related}
GCL addresses a fundamental challenge in robotics RL: the mismatch between simulated and real-world task distributions, which is often overlooked despite extensive work on improving simulator dynamics. We review relevant literature in CL and Unsupervised Environment Design (UED), highlighting how GCL improves upon existing approaches.

CL in RL aims to improve learning efficiency by progressively increasing simulated task complexity \cite{narvekar2020curriculum}. Recent work has focused on automatic curriculum generation \cite{zhang2020automatic}, where the curriculum is dynamically adapted based on the agent's performance in simulation. Such an adapted curriculum improves efficiency, but it doesn't have a mechanism to ensure that the generated tasks can represent the real world, where a physical robot will eventually be deployed~\cite{zhao2020sim}.

UED has emerged as a promising approach for automatically generating training tasks and adapting curricula in RL~\cite{parker2022evolving}. UED methods aim to reduce the need for manually designed tasks in simulation, a key challenge in robotic RL. While traditional UED methods like Domain Randomization~\cite{tobin2017domain} and minimax approaches~\cite{pinto2017robust,li2019robust} have shown effectiveness in simulated environments, they face challenges when applied to real-world robotics problems where data is scarce and expensive to obtain.

At the intersection of CL and UED, adaptive-teacher UED methods have shown promise in improving zero-shot generalization. PAIRED~\cite{dennis2020emergent} introduced a regret-based approach to UED, using an adversarial game to generate increasingly complex environments. 
CLUTR~\cite{azad2023clutr} improved upon PAIRED by introducing unsupervised representation learning to UED, replacing the explicit task generator with a learned latent space using Variational Autoencoders~\cite{kingma2013auto}, which allowed for more efficient task generation. 
However, CLUTR's approach is still limited in several key respects:
First, CLUTR's teacher agent lacks observation of the student's performance history, limiting its ability to adapt to the student's learning progress.
Second, because the teacher generates tasks using a stateless, multi-armed bandit algorithm, it lacks the capability to model the task space thoroughly, limiting its ability to generate complex tasks suitable for robotics scenarios.
Third, CLUTR operates solely in simulation and does not consider the challenges associated with sim-to-real transfer, particularly the mismatch between simulated and real-world task distributions.

GCL addresses these limitations by introducing a framework that grounds the curriculum in real-world data, ensuring that the learning process is directly applicable to real-world environments. 
GCL improves upon existing methods in three key respects: 
First, unlike simple RL test domains like Grid World, Car Racing, or video games~\cite{shao2019survey}, a key difference in robotic RL is that robots need to be eventually deployed in the real world after training in simulation. Therefore, GCL grounds the simulated learning process in real-world data, i.e., \emph{simulation realism}; 
Second, GCL grounds the task generation on the previous task sequences, i.e., \emph{task awareness}, and enables the teacher agent to manipulate the task space effectively; 
Third, GCL grounds the curriculum on monitoring of \emph{student performance} and allows every task to be catered to the student's latest capabilities (Fig.~\ref{fig:overview}).

\section{Approach}
\label{sec:approach}

\begin{figure}[t]
    \centering
    \includegraphics[width=\columnwidth]{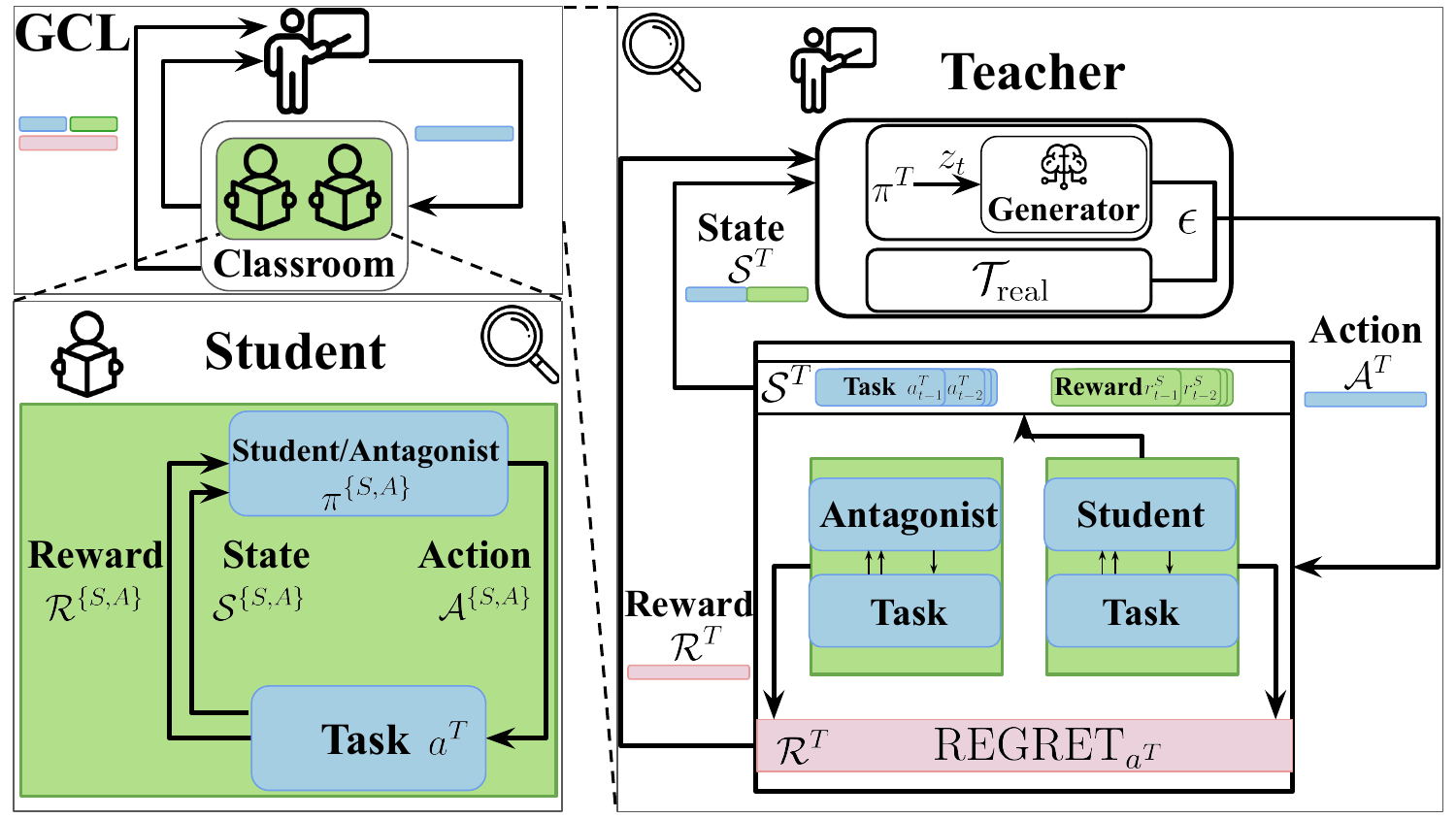}
    \caption{Overview of the Dual-Agent GCL Framework (top left): student POMDP (bottom left) and teacher MDP (right).}
    \label{fig:gcl_approach}
    \vspace{-15pt}
\end{figure}

We introduce GCL, a Dual-Agent framework for adaptive curriculum learning in robotics with limited real-world data. GCL consists of two interacting processes (Fig.~\ref{fig:gcl_approach}):
\begin{itemize}
    \item  A Partially Observable Markov Decision Process (POMDP) for the student agent ($\cdot^S$) learning the task.
    \item A fully informed Markov Decision Process (MDP) for the teacher agent ($\cdot^T$) generating a curriculum of tasks.
\end{itemize}

\subsection{Dual-Agent (PO)MDP}
\label{subsec:problem_formulation}

\subsubsection{Student Agent POMDP}
The student agent operates in a POMDP defined as a tuple $\mathcal{M}^S = \langle \mathcal{S}^S, \mathcal{A}^S, \mathcal{O}^S, \mathcal{T}^S, \Omega^S, \mathcal{R}^S, \gamma^S \rangle$, where:
\begin{itemize}
    \item $\mathcal{S}^S$ is the state space of the robotic task,
    \item $\mathcal{A}^S$ is the robot action space,
    \item $\mathcal{O}^S$ is the robot observation space,
    \item $\mathcal{T}^S: \mathcal{S}^S \times \mathcal{A}^S \rightarrow \mathcal{S}^S$ is the POMDP transition function,
    \item $\Omega^S: \mathcal{S}^S \times \mathcal{A}^S \rightarrow \mathcal{O}^S$ is the robot observation function,
    \item $\mathcal{R}^S: \mathcal{S}^S \times \mathcal{A}^S \times \mathcal{S}^S \rightarrow \mathbb{R}$ is the robot reward function based on task execution performance, and
    \item $\gamma^S \in [0, 1]$ is the student POMDP's discount factor.
\end{itemize}
The Student agent's goal is to learn a policy $\pi^S: \mathcal{O} \rightarrow \mathcal{A}$ that maximizes the expected cumulative reward in the partially observable task environment generated by the teacher.
\subsubsection{Teacher Agent MDP}
In contrast to the student's POMDP, the teacher agent operates in an MDP defined as $\mathcal{M}^T = \langle \mathcal{S}^T, \mathcal{A}^T, \mathcal{T}^T, \mathcal{R}^T, \gamma^T \rangle$, where:
\begin{itemize}
    \item $\mathcal{S}^T$ is the teacher state space, consisting of the comprehensive history of tasks and student performances,
    \item $\mathcal{A}^T$ is the teacher action space representing all possible tasks that can be assigned to the student,
    \item $\mathcal{T}^T: \mathcal{S}^T \times \mathcal{A}^T \rightarrow \mathcal{S}^T$ is the MDP transition function,
    \item $\mathcal{R}^T: \mathcal{S}^T \times \mathcal{A}^T \times \mathcal{S}^T \rightarrow \mathbb{R}$ is the teacher reward function based on student performance, and
    \item $\gamma^T \in [0, 1]$ is the discount factor for the teacher's MDP.
\end{itemize}
$s_t^T \in \mathcal{S}^T$ at time $t$ is defined as $s_t^T = \{(a_i^T, r_i^S)\}_{i=0}^{t-1}$, where $a_i^T \in \mathcal{A}^T$ is the $i$ th task assigned by the teacher and $r_i^S$ is the student's performance (reward) for task $i$.

\subsection{Grounded Curriculum Learning (GCL)}
\label{subsec:gcl}

GCL employs a hierarchical structure where a fully-informed teacher agent guides the learning process of a student agent, resembling a classroom setting (Fig.~\ref{fig:gcl_approach} top left). In this metaphorical classroom, the teacher (Fig.~\ref{fig:gcl_approach} right) oversees the learning environment (tasks in the curriculum) and monitors the performance of the student and an antagonist agent (Fig.~\ref{fig:gcl_approach} bottom left). This design reflects real-life educational scenarios, where teachers have comprehensive knowledge of both the curriculum and student progress. Leveraging this informed perspective within the classroom and dual-agent framework, GCL comprises five main components that work together to create an effective and adaptive curriculum:

\subsubsection{Task Representation via Latent Generative Model}
\label{subsubsec:latent_model}
We employ a Variational Autoencoder (VAE) to learn a compact latent space $\mathcal{Z}$ of robotic tasks, trained on a limited set of real-world tasks $\mathcal{T}_{\text{real}}$ and therefore grounded in the real world. The VAE consists of an encoder and a decoder, which learn to compress and reconstruct task environments efficiently.
This model enables the teacher to generate diverse, realistic tasks by sampling from the learned latent space $\mathcal{Z}$, bridging the gap between limited real-world data and the need for varied training scenarios. By learning a continuous task representation, the VAE allows smooth interpolation between known tasks and the generation of novel, yet realistic, ones for the student agent to learn from.

\subsubsection{Student and Antagonist Agents}
\label{subsubsec:student_agent}
The student agent learns to perform a task using a reinforcement learning algorithm (e.g., PPO \cite{schulman2017proximal}) in the partially observable environment generated by the teacher. Its objective is to maximize the expected cumulative reward:
\begin{equation}
J^S(\pi^S_{\theta^S}) = \mathbb{E}_{\tau^S \sim \pi^S_{\theta^S}}\left[\sum_{t=0}^{T} {(\gamma^S)}^t r^S_t\right],
\label{eqn::J^S}
\end{equation}
where $\tau^S$ is a trajectory sampled from the student policy $\pi^S_{\theta^S}$, parameterized by $\theta^S$.
To guide curriculum generation and evaluate the student's progress, we introduce an antagonist agent, following the flexible regret setting from PAIRED~\cite{dennis2020emergent}. The antagonist is trained with the same observability and hyperparameters as the student, sharing the same objective function (Eqn.~\eqref{eqn::J^S}), with $J^A$ and $\pi^A_{\theta_A}$ as the antagonist's objective and policy respectively.

\subsubsection{Teacher Agent}
\label{subsubsec:teacher_agent}

The teacher agent enables GCL's two grounding aspects:  student performance and task awareness.  

To ground in student performance, the teacher maintains a comprehensive tracking of the student's performance across diverse tasks through the student's historical performance, i.e., $\{r_i^S)\}_{i=0}^{t-1}$, as part of the teacher state $s_t^T$. By incorporating performance history into its state, the teacher can adapt the curriculum based on the student's learning progress.

To ground with task awareness, the teacher maintains a deep understanding of the task space, ensuring continual relevance to past and real-world tasks. This awareness is facilitated by both the historical tasks, i.e.,  $\{a_i^T\}_{i=0}^{t-1}$, in the teacher state $s_t^T$, as well as the latent space $\mathcal{Z}$ learned through the VAE from real-world tasks. According to the task history, the teacher generates new tasks for the student by sampling from this latent space using the VAE decoder $G: \mathcal{Z} \rightarrow \mathcal{A}^T$, which maps latent vectors to concrete tasks.

After grounding in terms of student performance and task awareness, the teacher's objective is to maximize the expected cumulative regret:

\begin{equation}
    J^T(\pi^T_{\theta^T}) = \mathbb{E}_{\tau^T \sim \pi^T_{\theta^T}}\left[\sum_{t=0}^{T} {(\gamma^T)}^t \textsc{Regret}_{a^T_t}(\pi^S_{\theta^S}, \pi^A_{\theta_A})\right], \nonumber
\end{equation}
where:
\begin{itemize}
    \item $\tau^T = (s_0^T, a_0^T, s_1^T, a_1^T, ..., s^T_t)$ is a trajectory in the teacher's MDP, 
    \item $\pi^T_{\theta^T}$ is the teacher policy, parameterized by $\theta^T$, 
    \item $\textsc{Regret}_{a^T_t}(\pi^S_{\theta^S}, \pi^A_{\theta_A}) = V_{a^T_t}(\pi^A_{\theta_A}) - V_{a^T_t}(\pi^S_{\theta^S})$ is the flexible regret for task $a^T_t$, generated by the teacher at time $t$, and
    \item $V_{a^T_t}(\cdot)$ is the value function (expected discounted return) of a policy when executing task $a^T_t$.
\end{itemize}
\begin{algorithm}[tbp]
\small
\caption{Grounded Curriculum Learning (GCL)}
\label{alg:gcl}
\begin{algorithmic}[1]
\STATE \textbf{Input:} VAE decoder $G$, initial parameters $\theta^S$, $\theta^A$, $\theta^T$, learning rates $\eta^S$, $\eta^A$, $\eta^T$, real-world task set $\mathcal{T}_{\text{real}}$, and grounding probability $\epsilon$
\STATE \textbf{Output:} Trained policies $\pi^S_{\theta^S}$, $\pi^A_{\theta^A}$, and $\pi^T_{\theta^T}$
\STATE Pretrain $G$ with available real-world tasks $\mathcal{T}_{\text{real}}$
\STATE Initialize $\pi^S_{\theta^S}$, $\pi^A_{\theta^A}$, $\pi^T_{\theta^T}$, and $s_0^T = \{\}$
\STATE $t \gets 0$
\WHILE{not converged}

    \STATE ${a_t^T} \gets \begin{cases}
        \text{sample from } \mathcal{T}_{\text{real}}, & \text{with probability } \epsilon, \\
        \pi^T_{\theta^T}(s_t^T), & \text{with probability } 1-\epsilon,
    \end{cases}$
    \STATE Collect student trajectory $\tau^S = \{(o_t^S, {a_t^T}^S, r_t^S)\}_{t=0}^T$ in task ${a_t^T}$ and compute rewards $r^S_{{a_t^T}}$ 
    \STATE Collect antagonist trajectory $\tau^A = \{(o_t^A, {a_t^T}^A, r_t^A)\}_{t=0}^T$ in task ${a_t^T}$

    \STATE Compute regret $\textsc{Regret}_{{a_t^T}} \gets V_{{a_t^T}}(\pi^A_{\theta^A}) - V_{{a_t^T}}(\pi^S_{\theta^S})$
    \STATE Update $s_{t+1}^T \gets s_t^T \cup \{({a_t^T}, r^S_{{a_t^T}})\}$
    \STATE $\pi^S \gets \pi^S + \alpha^S \nabla_{\pi^S} J^S(\pi^S)$
    \STATE $\pi^A \gets \pi^A + \alpha^A \nabla_{\pi^A} J^A(\pi^A)$
    \STATE $\pi^T \gets \pi^T + \alpha^T \nabla_{\pi^T} J^T(\pi^T)$
 
    \STATE $t \gets t + 1$
\ENDWHILE
\RETURN $\pi^S_{\theta^S}$, $\pi^A_{\theta^A}$, and $\pi^T_{\theta^T}$
\end{algorithmic}

\end{algorithm}
\subsubsection{Grounding Simulated Tasks in the Real World}
\label{subsec:balancing_tasks}

GCL implements a novel mechanism for balancing real-world and simulated tasks. The teacher agent employs a probabilistic approach to task selection, sampling task ${a_t^T}$ between generated tasks and real-world tasks:
\begin{equation}
    {a_t^T} = \begin{cases}
        \text{sample from } \mathcal{T}_{\text{real}}, & \text{with probability } \epsilon, \\
        \pi^T_{\theta^T}(s_t^T), & \text{with probability } 1-\epsilon, \nonumber
    \end{cases}
\end{equation}
where $\epsilon \in [0,1]$ is a hyperparameter controlling the balance between real and simulated tasks.
This approach ensures regular grounding in real-world scenarios while allowing for curriculum adaptation through generated tasks. By adjusting $\epsilon$, we can control the degree of grounding in real-world data, making the framework flexible to different learning scenarios and the availability of real-world data.

Algorithm \ref{alg:gcl} implements the hierarchical structure of GCL, encapsulating its key components. Using a pre-trained VAE~\cite{prakash2019use} (line 3), the fully informed teacher agent alternates between sampling real-world tasks and generating new ones (line 7). 

The partially observable student and antagonist agents collect trajectories in the selected task environment $a^T_t$ (lines 8-9). The teacher's state is updated with each new task-regret pair (lines 10-11), enabling curriculum adaptation based on the full history of tasks and performances. All policies are updated using their respective objective functions (lines 12-14). 
The process continues until convergence.

\section{Experiments}
\label{sec:experiments}

We evaluated GCL on The Benchmark for Autonomous Robot Navigation (BARN) Challenge \cite{perille2020benchmarking, xu2023benchmarking, xu2021machine, xiao2022autonomous, xiao2023autonomous, xiao2024autonomous}, a standardized testbed for SOTA navigation systems designed to push the boundaries of performance in challenging and highly constrained environments. The objective is to navigate a robot from a predefined start to a goal location as quickly as possible without collisions. Focusing on real-world autonomous navigation, BARN features an environment generator capable of producing a wide range of navigation tasks, from easy open spaces to difficult highly constrained ones.

\subsection{Experimental Setup}
\label{subsec:exp_setup}

We implement The BARN Challenge in NVIDIA's IsaacGym simulator \cite{makoviychuk2021isaac}, utilizing 128 parallel environments. This setup significantly accelerates training and allows efficient exploration of the task space for curriculum learning. Considering the lack of large-scale real-world scenarios, in our experiments, the BARN environments from the generator serve as a surrogate for the real world, where robots will be eventually deployed after training (in contrast to simulated environments created by the teacher agent during training). 
\subsubsection{Student Agent POMDP}
$\mathcal{M}^S = \langle \mathcal{S}^S, \mathcal{A}^S, \mathcal{O}^S, \mathcal{T}^S, \Omega^S, \mathcal{R}^S, \gamma^S \rangle$ is implemented as a navigation task in our experiments. 
$\mathcal{S}^S$ includes the robot's position and orientation;
$\mathcal{A}^S$ comprises continuous linear and angular velocities;
$\mathcal{O}^S$ includes 270° field-of-view, 720-dimensional LiDAR scans and the relative goal orientation;
and $\mathcal{R}^S$ encourages progress towards the goal while penalizing collisions and excessive time.

\subsubsection{Teacher Agent MDP}
In $\mathcal{M}^T = \langle \mathcal{S}^T, \mathcal{A}^T, \mathcal{T}^T, \mathcal{R}^T, \gamma^T \rangle$, 
$\mathcal{S}^T$ consists of the history of tasks and student performances, i.e., $s_t^T = \{(a_i^T, r_i^S)\}_{i=0}^{t-1}\in \mathcal{S}^T$. For simplicity, we set $i=t-1$ in our experiments and leave the study on the effect of history length as future work;
$\mathcal{A}^T$ is the latent space of task representation; and $\mathcal{R}^T$ is based on the regret between the student and the antagonist.

\subsubsection{Hyperparameters}
Table \ref{tab:hyperparameters} summarizes the key hyperparameters used in our experiments.
\begin{table}[tbp]
\centering

\caption{Hyperparameters for GCL.}
\label{tab:hyperparameters}
\begin{tabular}{lll}
\specialrule{1.2pt}{0pt}{0pt} 

GCL Parameter & Value & \\
\midrule
Parallel Environments & 128 & \\
Latent Task Dimension & 32 & \\
Training Epochs & 5000 & \\
\midrule
RL Parameter & Teacher & Student \\
\midrule
Learning Rate & 1e-4 & 3e-4 \\
PPO Epoch & 10 & 5 \\
Discount Factor & 0.99 & 0.99 \\
\specialrule{1.2pt}{0pt}{0pt} 

\end{tabular}
\vspace{-10pt}
\end{table}

\subsection{Methods and Evaluation Metrics}
\label{subsec:methods_and_metrics}
\begin{table}[tbp]
\centering
\caption{Comparison of Methods Used in the Experiments.}
\label{tab:methods_comparison}

\begin{tabular}{llll}
\specialrule{1.2pt}{0pt}{0pt} 

Method & Task Environment & Curriculum & Teacher Agent \\
\midrule
Base RL & Real-World Tasks & None & N/A \\
Manual CL & Real-World Tasks & Hand-Designed & Manual \\
CLUTR & Simulated & Automatic & Stateless \\
\textbf{GCL} & \textbf{Real +Simulated} & \textbf{Automatic} & \textbf{Fully Informed}\\

\specialrule{1.2pt}{0pt}{0pt} 

\end{tabular}
\vspace{-10pt}
\end{table}
We compare GCL against three baseline approaches: Base RL, Manual RL~\cite{xu2024reinforcemen}, and CLUTR~\cite{azad2023clutr}. Table \ref{tab:methods_comparison} summarizes the key characteristics of GCL and these baseline methods. For the Manual RL approach, we construct an expert curriculum based on the shortest path length to traverse each environment. This metric serves as our heuristic for task difficulty, allowing us to create a hand-designed curriculum that progressively increases the complexity of navigation tasks.

For evaluation, we employ a comprehensive set of metrics to assess various aspects of navigation performance. These include Task Success rate, which measures the percentage of trials where the robot successfully reaches the goal position without collisions; Navigation Progress, which reflects the average proportion of the path completed before success or failure; average steps taken per successful task; average reward accumulated; and average speed of the robot during tasks. 
To ensure a fair evaluation, we utilize the BARN environment generator to create separate training (for grounding on simulation realism) and test sets. The environments are split in a 70/30 ratio for training and testing. This arrangement guarantees that the robot has never encountered the evaluation environments during the training phase, allowing us to assess the generalization capabilities of the learned policies.

\begin{table*}[t]
\centering
\setlength{\tabcolsep}{4pt} 
\small 
\begin{tabular*}{\textwidth}{@{\extracolsep{\fill}}lccccc@{}}
\specialrule{1.2pt}{0pt}{0pt} 
\textbf{Method} & \textbf{Task Success (\%) $\uparrow$} & \textbf{Navigation Progress (\%) $\uparrow$} & \textbf{Avg. Steps $\downarrow$} & \textbf{Avg. Reward $\uparrow$} & \textbf{Avg. Speed (m/s) $\uparrow$} \\
\midrule
Base RL & 76.16 ± 4.47 & 64.06 ± 2.38 & \textbf{36.41 ± 0.15} & 18.36 ± 0.84 & 1.97 ± 0.01 \\
Manual CL & 76.83 ± 5.02 & 66.17 ± 2.47 & 36.65 ± 0.22 & 19.19 ± 1.17 & \textbf{1.99 ± 0.00} \\
CLUTR & 76.67 ± 2.74 & 66.52 ± 4.23 & 36.70 ± 0.28 & 18.39 ± 0.65 & 1.97 ± 0.01 \\
\textbf{GCL} & \textbf{81.85 ± 2.51} & \textbf{68.89 ± 2.77} & 36.99 ± 0.57 & \textbf{19.45 ± 0.77} & 1.94 ± 0.01 \\
\specialrule{1.2pt}{0pt}{0pt} 
\end{tabular*}
\caption{Test Performance Comparison of Different Methods across Various Metrics (mean ± std) in The BARN Challenge.}
\label{tab:performance_comparison_plot}
\vspace{-15pt}
\end{table*}

\subsection{Main Results}
\label{subsec:main_results}

Table \ref{tab:performance_comparison_plot} presents a comparison of all four methods averaged over three runs. Upward and downward pointing arrows next to each metric indicate whether higher or lower values are better, respectively. 
GCL achieves the highest Task Success rate (81.85\%) and Navigation Progress (68.89\%), outperforming all other methods and demonstrating its superior ability to successfully complete navigation tasks and make significant progress even in failed attempts. GCL also accumulates the highest Avg. Reward (19.45).
In contrast, Base RL and Manual CL achieve the best Avg. Steps and Avg. Speed, reflecting a more aggressive navigation strategy, but this comes at the cost of increased and earlier failures during task execution.

\subsection{Ablation Studies}
\label{subsec:ablation_studies}
We conduct comprehensive ablation studies to analyze the contribution of each key component in GCL:
\begin{itemize}
    \item \texttt{GCL}: The complete GCL framework.
    
    \item \texttt{GCL w/o real}: GCL without real-world data grounding. We remove the real-world task selection, using only tasks generated by the teacher agent: ${a_t^T} = \pi^T_{\theta^T}(s_t^T)$.

    \item \texttt{GCL w/o task}: GCL without task awareness grounding. We replace the latent task representation $a_i^T$ with a random vector $\xi_i$ in the teacher's state: 
        $s_t^T = \{(\xi_i, r_i^S)\}_{i=0}^{t-1}$, where $\xi_i \sim \mathcal{N}(0, I)$, a zero-mean, identity-standard-deviation normal distribution.
    
    \item \texttt{GCL w/o performance}: GCL without performance history grounding. We replace the student's reward $r_i^S$ with a random scalar $\eta_i$ in the teacher's state: 
        $s_t^T = \{(a_i^T, \eta_i)\}_{i=0}^{t-1}$, where $\eta_i \sim \mathcal{U}(0, 1)$, a uniform distribution between $0$ and $1$. 
\end{itemize}

\begin{table}[tbp]
\centering
\renewcommand{\arraystretch}{1.2}
\begin{tabular}{lcc}
\specialrule{1.2pt}{0pt}{0pt} 

\textbf{Method} & \textbf{Success Rate (\%)} & \textbf{Performance Gain (\%)} \\
\hline
GCL w/o real & 76.36 & +7.19 \\
GCL w/o task & 79.86 & +2.49 \\
GCL w/o performance & 77.69 & +5.35 \\
\textbf{GCL} & \textbf{81.85} & -- \\

\specialrule{1.2pt}{0pt}{0pt} 

\end{tabular}
\caption{Success Rate Comparison of GCL and Its Ablated Variants in The BARN Challenge: Each variant removes one grounding component, demonstrating the importance of each aspect in the full GCL framework. The Performance Gain column shows the percentage improvement of complete GCL over each variant.}
\label{tab:performance_comparison_ablation_study}
\vspace{-10pt}
\end{table}
Table~\ref{tab:performance_comparison_ablation_study} presents the results of our ablation studies. The performance gains of the full GCL over its variants emphasize the critical importance of each component in the framework. Grounding in real-world data is the most crucial element, followed by performance history and task awareness. These components work synergistically to enhance GCL's effectiveness, leveraging the advantages of curriculum learning while maintaining a strong connection to real-world scenarios. These results highlight the importance of a holistic approach that combines real-world relevance with adaptive learning strategies based on task and performance understanding, leading to superior performance in complex navigation tasks.

\begin{figure*}[tbp]
\centering

\includegraphics[width=\textwidth]{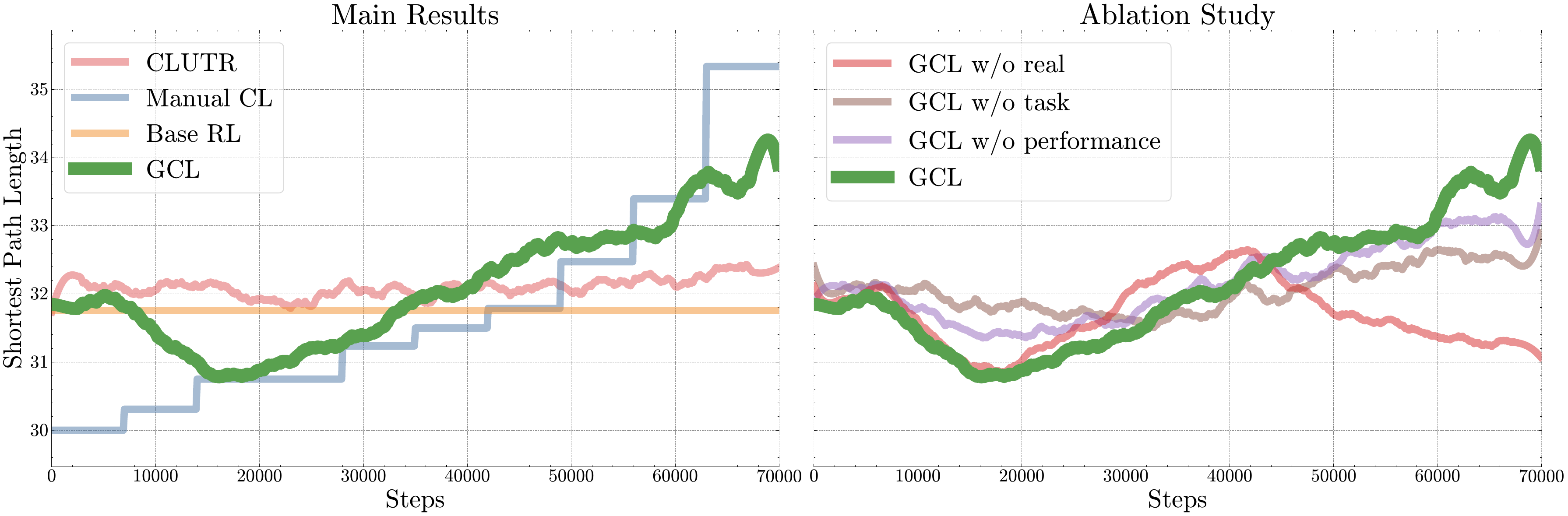}
\caption{Task Difficulty Adaptation during Curriculum Training for GCL and Comparative Methods.}
\vspace{-15pt}
\label{fig:difficulty_progression}
\end{figure*}

\subsection{Curriculum Progression}
We present and discuss the curriculum progression enabled by different methods to investigate GCL's capability to autonomously create appropriate tasks to facilitate learning.  
\subsubsection{Curriculum Progression based on Heuristic Difficulty}
Fig. \ref{fig:difficulty_progression} illustrates the progression of task difficulty during curriculum training. The x-axis represents training steps, while the y-axis depicts the shortest path length, our heuristic metric for task difficulty. Higher values indicate more difficult and complex tasks, typically involving longer and more tortuous paths. It's crucial to acknowledge that this difficulty metric, based on human intuition, may not fully capture the true difficulty experienced by the student agent.
\subsubsection{Main Method Comparison }
The Fig. \ref{fig:difficulty_progression} (left) compares four methods: Baseline RL, Manual CL, CLUTR, and our proposed GCL.
Baseline RL maintains a constant difficulty level and does not adjust to the agent's evolving capabilities in complex scenarios.
Manual CL, designed by experts, shows structured progression but relies heavily on task-specific knowledge, limiting its applicability in novel or rapidly evolving domains.
CLUTR, despite automatic generation, shows limited variation in task difficulty over time, possibly due to insufficient consideration of student performance and task space understanding.
In contrast, GCL demonstrates dynamic adaptation of task difficulty throughout training. Its fluctuating curve indicates responsiveness to the agent's current capabilities, autonomously discovering effective learning progressions without relying on task-specific expert knowledge.

\subsubsection{Ablation Study Comparison}
The Fig. \ref{fig:difficulty_progression} (right) presents our ablation study, comparing GCL variants. Although all variants generate a curriculum, the full GCL model shows the most adaptive approach, adjusting task difficulty based on the agent's learning progress.
Notably, while \texttt{GCL w/o real} generates a curriculum, its poor performance highlights that curriculum generation alone is insufficient without grounding in real-world data. This finding underscores the critical importance of incorporating real-world information in curriculum design for robotic learning tasks.
\begin{figure}[tbp]
\centering

\includegraphics[width=\columnwidth]{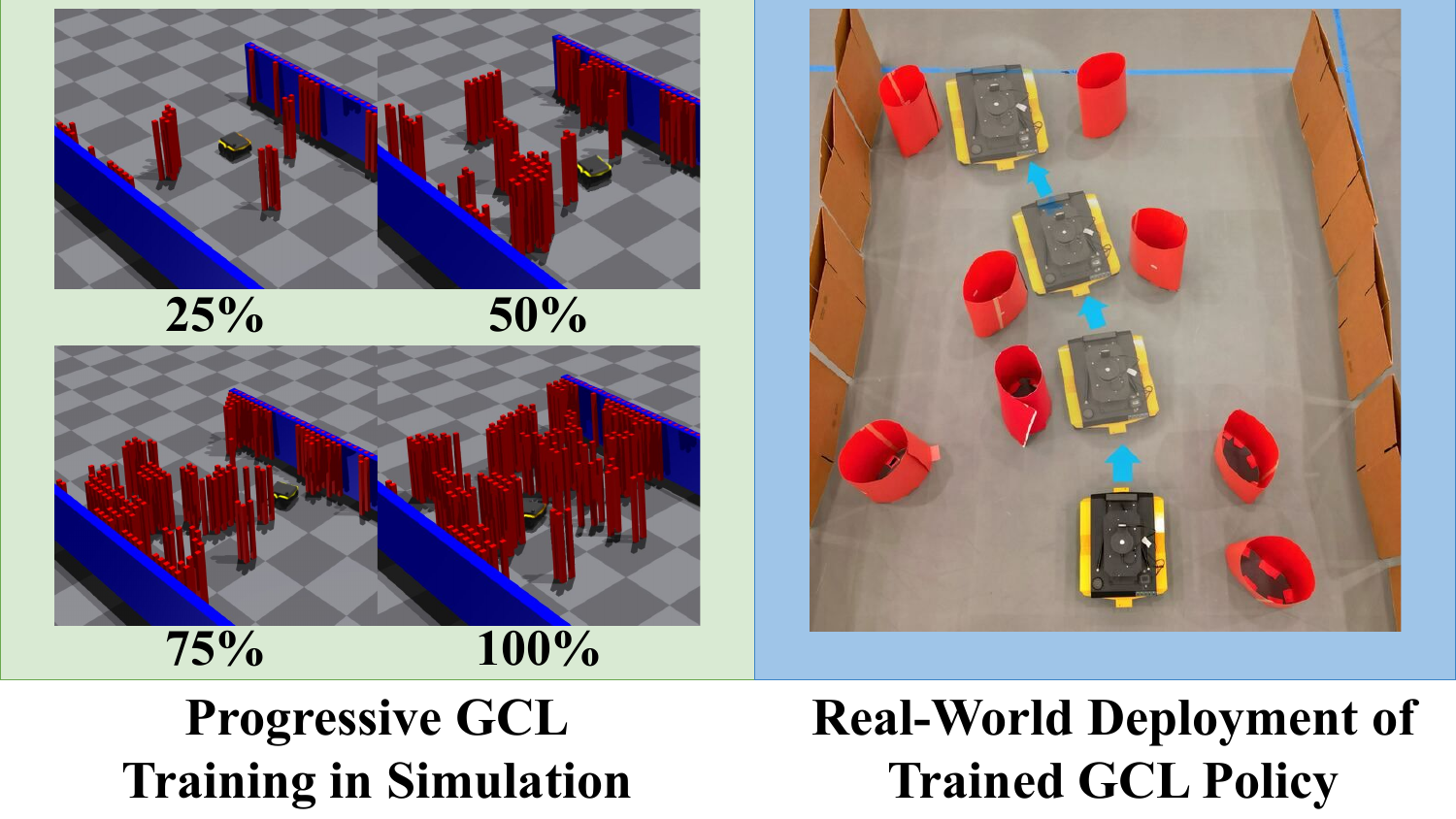}
\caption{GCL from Simulation to Reality. Left: Progressive stages of GCL training in simulation; Right: Deployment of the trained GCL policy in the physical real world.}
\label{fig:environment_visualization}
\vspace{-18pt}
\end{figure}

\subsubsection{Environment Visualization}

Fig.~\ref{fig:environment_visualization} illustrates GCL's progression from simulation training to real-world deployment. The left side presents environments generated by GCL at four stages of training (25\%, 50\%, 75\%, and 100\%), demonstrating the framework's capacity to develop an adaptive curriculum. In the initial stage (25\%), GCL generates environments with minimal obstacles, facilitating the acquisition of fundamental navigation skills. As training progresses through 50\% and 75\%, the environments exhibit increasing complexity, introducing more obstacles and intricate pathways. At the final stage (100\%), GCL produces environments that test advanced navigation capabilities.
The right side of the figure shows the robot navigating a real-world BARN environment using the trained GCL policy. The robot can maneuver through the environment, skillfully avoiding obstacles and ultimately crossing the blue goal line. The real-world deployment demonstrates the robot's capacity to apply simulation-learned skills to navigate complex, real-world settings, indicating the practical applicability of the GCL approach. Considering our usage of the simulated BARN environments from the BARN generator as a surrogate for the real world (in contrast to fully synthetic environments produced by the teacher agent), deploying the learned policy in the physical real world introduces another sim-to-real gap (in addition to the gap due to environment distribution), e.g., due to different physics and robot models. So the GCL policy also fails from time to time in the physical real world. 

\subsubsection{Discussions}
The curriculum progression analysis reveals a paradigm shift from rigid, predetermined learning structures to adaptive, autonomous curriculum generation. GCL's performance illustrates the potential of integrating real-world grounding with flexible, agent-responsive learning strategies.
Our comparative analysis highlights the balance between structure and flexibility in GCL: While structured approaches provide clear learning paths, they may inadvertently limit the exploration of other potentially more efficient learning strategies; GCL's approach allows for the discovery of unexpected yet effective learning pathways, potentially leading to more robust and generalizable robotic skills.
Furthermore, the results emphasize the importance of bridging the gap between simulated and real-world learning in robotics. As we continue to push the boundaries of robotic capabilities, approaches like GCL that can autonomously generate and adapt curricula may become increasingly crucial, offering a paradigm for developing more versatile and efficient robotic systems capable of tackling the complexities of real-world environments.

\section{Conclusions and Future Work}
\label{sec::conclusions}

This paper introduces Grounded Curriculum Learning (GCL), a novel framework that enhances real-world reinforcement learning in robotics. GCL improves curriculum learning by aligning the simulated task distribution with real-world tasks while considering both task sequences and the robot’s past performance. Our experiments on the BARN navigation dataset demonstrate GCL's effectiveness, achieving a 6.8\% and 6.5\% higher success rate compared to SOTA methods and a manually designed curriculum. GCL's balance between structured and flexible learning ensures that the learned skills are both efficiently acquired and applicable to real-world scenarios.  
Our ablation studies demonstrate that each component of GCL is important: the curriculum (grounded in task awareness and performance history) guides efficient learning, while real-world task grounding is crucial for maintaining relevance to the target domain.

An interesting direction for future work is to extend GCL to a broader range of robotic tasks beyond navigation~\cite{xiao2022motion}, such as manipulation~\cite{haarnoja2018composable, kalashnikov2018scalable} and multi-agent systems~\cite{limbu2023team, limbu2024scaling, zhouIROS2024, liu2021team}. Another promising avenue is investigating methods for more effective latent space manipulation by the teacher agent, potentially leading to improved task generation and curriculum design. Additionally, exploring GCL's potential in transfer and lifelong learning scenarios~\cite{liu2021lifelong} with a pre-trained teacher agent can enable robots to adapt more quickly to new tasks or environments. Lastly, instead of end-to-end learning~\cite{pfeiffer2017perception, datar2024toward}, more structured learning approaches for robotics,  such as learning planner parameters~\cite{xiao2022appl, xiao2020appld, wang2021appli, wang2021apple, xu2021applr, das2024motion}, cost functions~\cite{xiao2022learning}, kinodynamic models~\cite{xiao2021learning, karnan2022vi, atreya2022high, datar2024terrain, pokhrel2024cahsor}, trajectory generation~\cite{liang2024mtg, liang2024dtg}, and local planners~\cite{xiao2021toward, xiao2021agile, wang2021agile, raj2024rethinking} may be able to further improve GCL's efficiency and generalizability.

\clearpage
\bibliographystyle{IEEEtran}
\bibliography{IEEEabrv,references}

\end{document}